\documentclass[sigconf]{acmart}
\AtBeginDocument{%
  }
\usepackage{multirow}
\copyrightyear{2026}
\acmYear{2026}
\setcopyright{cc}
\setcctype{by}
\acmConference[MM '26]{Proceedings of the 34th ACM International Conference on Multimedia}{November 10--14, 2026}{Rio de Janeiro, Brazil}
\acmBooktitle{Proceedings of the 34th ACM International Conference on Multimedia (MM '26), November 10--14, 2026, Rio de Janeiro, Brazil}
\acmDOI{10.1145/3767308.3835813}
\acmISBN{979-8-4007-2213-4/2026/11}

\acmSubmissionID{5038}

\begin{document}

\title{Learning Text-to-Motion from Partially Observed Web Motions via Part-Aware Masked Modeling}

\author{Boyuan Li}
\affiliation{%
  \institution{Renmin University of China}
  \city{Beijing}
  \country{China}
}
\email{liboyuan@ruc.edu.cn}

\author{Sipeng Zheng}
\affiliation{%
  \institution{BeingBeyond}
  \city{Beijing}
  \country{China}
}
\email{spzheng@beingbeyond.com}

\author{Bin Cao}
\affiliation{%
  \institution{Institute of Automation, Chinese Academy of Sciences}
  \city{Beijing}
  \country{China}
}
\email{caobin2023@ia.ac.cn}

\author{Ruihua Song}
\authornote{Corresponding author.}
\affiliation{%
  \institution{Renmin University of China}
  \city{Beijing}
  \country{China}
}
\email{rsong@ruc.edu.cn}

\author{Zongqing Lu}
\affiliation{%
  \institution{Peking University}
  \city{Beijing}
  \country{China}
}
\email{zongqing.lu@pku.edu.cn}

\renewcommand{\shortauthors}{Boyuan Li, Sipeng Zheng, Bin Cao, Ruihua Song, and Zongqing Lu}

\begin{abstract}
Internet videos are a promising complement to motion-capture data for text-to-motion generation, but most web shots do not contain a fully visible body. Existing pipelines either keep all recovered joints as if they were equally reliable targets or discard incomplete clips altogether, yet both clip-level policies are mismatched to web motion with mixed-quality body parts. We study this setting as part-level credible supervision, where observed parts provide reliable learning targets while unobserved parts should not directly define the loss. We propose RoPAR, a part-aware text-to-motion framework that uses only the credible body parts as direct supervision and infers the remaining parts from text and visible context. Specifically, we divide the body into five kinematic parts, estimate part credibility from web videos, learn a latent motion space from the credible part motions, and use masked generation to reconstruct the full motion sequence. We also build K700-M, a dataset of 198k motion-text pairs mined from Kinetics-700 for studying this setting. On K700-M, RoPAR outperforms both clip-level alternatives under matched \textit{Keep-All} and \textit{Discard} protocols, reducing the best baseline FID by 19.1\% and improving the best baseline R@1 by 20.3\% in the scale-preserving \textit{Keep-All} setting. These results suggest that, for web-recovered motion, keeping partially useful clips and restricting supervision to image-supported body parts is more effective than clip-level filtering.
\end{abstract}



\maketitle

\begin{figure}[!h]
\centering
\includegraphics[width=0.9\linewidth]{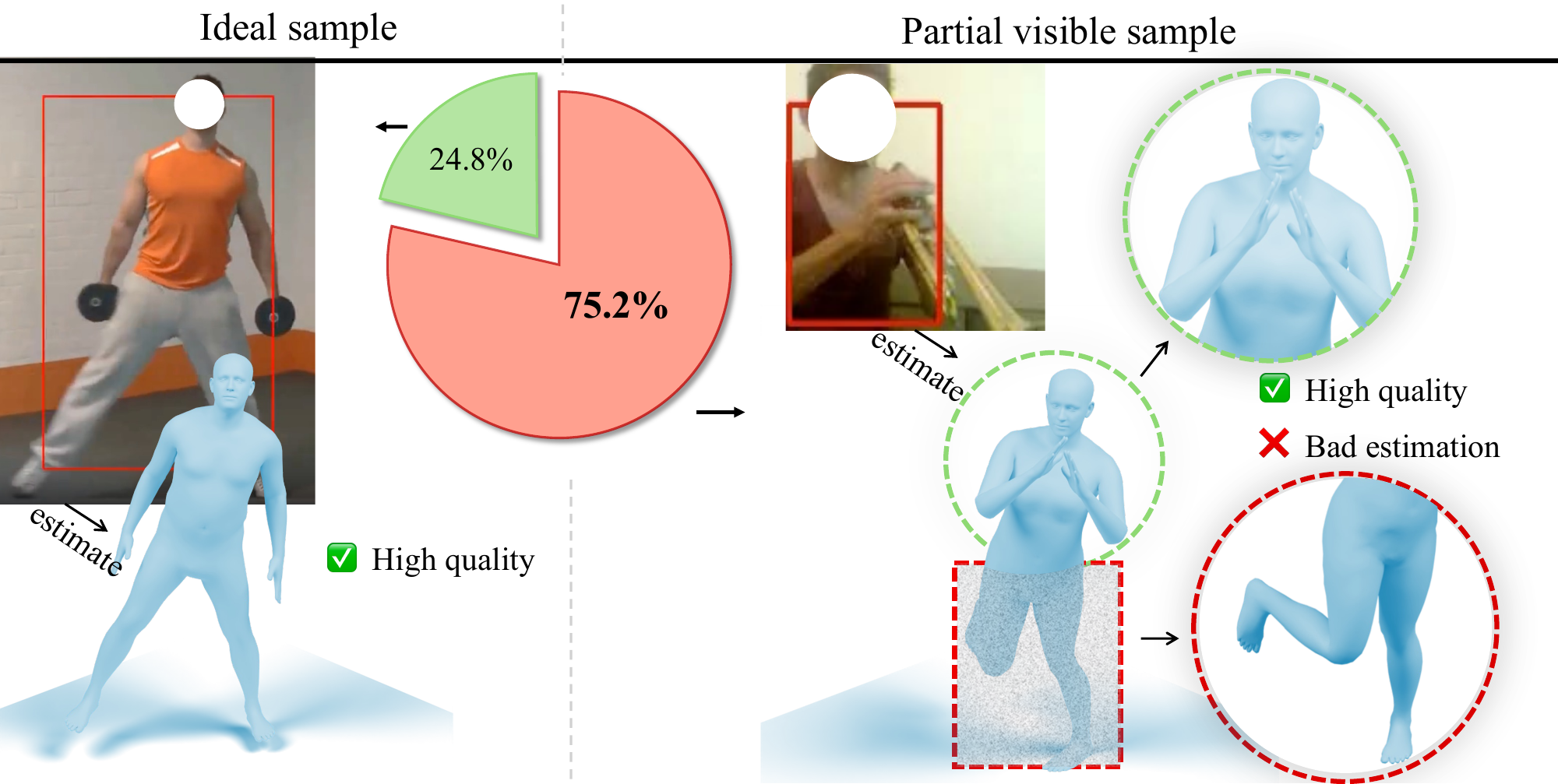}
\caption{Web-recovered motion is usually only partially observed. Our pilot analysis on Kinetics-700 shows that only 24\% of frames contain a fully visible body; in the remaining frames, at least one body part must be hallucinated from incomplete visual evidence.}
\label{fig:visibility_pie}
\end{figure}

\section{Introduction}
\label{sec:intro}

Text-to-motion (T2M) models need large motion-text corpora, yet high-quality motion capture data is still limited in scale, scene diversity, and action coverage~\cite{humanml3d,amass}. Web video can help: it contains open-vocabulary actions, object interactions, and camera viewpoints that curated MoCap datasets rarely capture. Recent efforts such as Motion-X and Being-M0 scale T2M with web-recovered motion~\cite{motion-x,being-m0,wham,whac}. Our focus is narrower. Once 3D motion is recovered from raw videos with Human Mesh Recovery (HMR) pipelines, the resulting supervision is often only partially credible because not every body part is directly supported by image evidence.

Web videos seldom show the full body for an entire clip. People are cropped, self-occluded, or blocked by objects and other people. HMR still outputs a complete motion sequence in these cases by filling in missing joints with learned priors. The result often looks plausible, but plausibility is not the same as observability: some joints are supported by image evidence, while others are guesses. Treating them as equally reliable targets creates a mismatch between the source video and the supervision signal.

Our pilot analysis on Kinetics-700~\cite{k700}, summarized in Fig.~\ref{fig:visibility_pie}, shows that this issue is common. Only 24\% of frames contain a fully visible body. In the remaining 76\%, at least one part has to be inferred from incomplete evidence. At the sequence level, more than 70\% of K700-M clips contain at least one non-visible body part. Web-recovered motion is therefore usually only partially observed.

One obvious fix is to drop incomplete clips. The problem is that this would remove most web data, including many clips that still contain useful local motion. A talking-head video may still provide reliable upper-body dynamics, and a cooking clip may still reveal informative arm motion. Many web clips remain useful at the part level even when the whole body is not. The challenge is to learn from the observed parts without forcing the model to imitate unsupported ones.

\begin{figure}[t]
\centering
\includegraphics[width=0.9\linewidth]{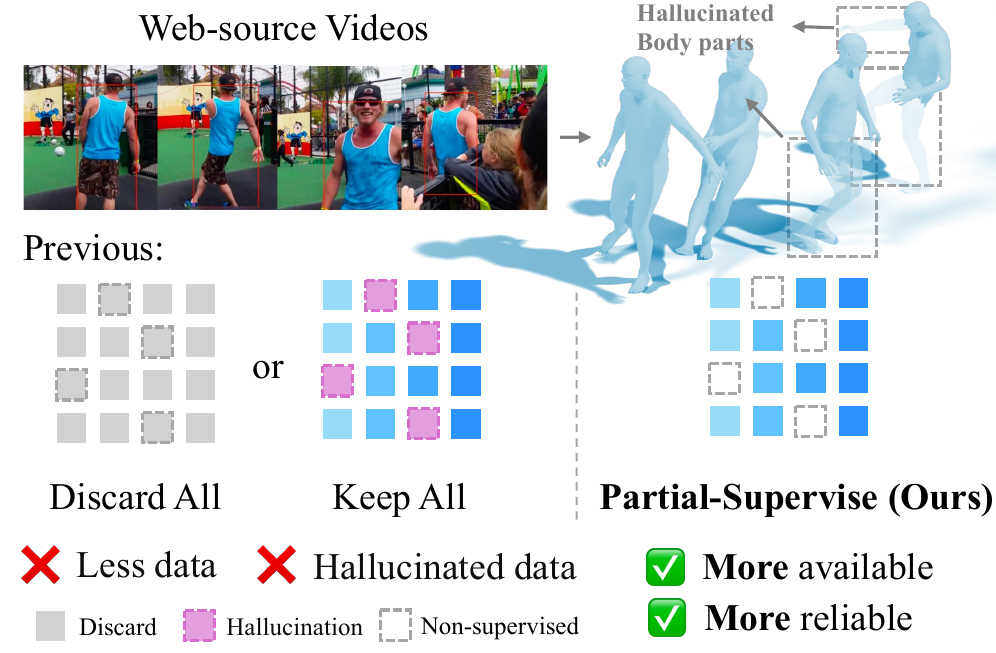}
    \caption{Compared with the two standard clip-level policies for web-recovered motion, RoPAR keeps the clips but supervises only the visible parts. This avoids both wasting useful partial evidence under \textit{Discard} and learning directly from unsupported HMR completions under \textit{Keep-All}.}
\label{fig:motivation}
\end{figure}

Current T2M pipelines, illustrated in Fig.~\ref{fig:motivation}, mostly fall back on two clip-level policies for web-recovered motion. \textit{Keep-All} treats the full recovered body as supervision, even when some parts were never observed. \textit{Discard} removes any clip with a non-visible body part, which means throwing away most of the source pool. Both policies fit web motion poorly because reliability varies within clips and across the dataset. The supervision rule has to work at the part level.

RoPAR is built around a simple learning principle: web-mined T2M should selectively trust the body parts that are supported by image evidence. Reconstructed joints should not all be treated as ground truth. We decompose the body into five kinematic parts, estimate which parts are credible in each frame from 2D pose confidence, and use only those parts as direct supervision. Non-credible parts are masked out of the target but still remain as missing context. A part-aware latent model learns from the credible motions, and a masked generator infers the missing parts from text and the remaining visible context before a diffusion head refines the full-body sequence. The result is a model that turns partial visibility into a supervision signal and reduces label noise.

To study this setting at scale, we build K700-M from Kinetics-700~\cite{k700}. K700-M contains 198,627 motion-text pairs recovered from real web videos and is intentionally harder than MoCap-based datasets because truncation, occlusion, and camera bias are common. We use it as a dataset and evaluation testbed for T2M under partial observability. It is not meant to be a universal ranking benchmark. On K700-M, RoPAR outperforms strong baselines under both standard web-data policies: keeping the full recovered source pool or discarding partially observed clips. We also evaluate on HumanML3D~\cite{humanml3d} as a fully credible control to check whether the same architecture still works when partial credibility is not an issue. Put differently, RoPAR is aimed at the supervision problem inside web-scale T2M. Other work expands coverage and capacity. We focus on what the model should trust once web-recovered motion is only partly observed.

Our contributions are threefold:
\begin{itemize}
\item We formulate web-mined T2M under partial observability as a problem of part-level credible supervision, and show why the two standard clip-level web-data policies, \textit{Keep-All} and \textit{Discard}, are mismatched to this regime.
\item We propose RoPAR, a part-aware masked modeling framework that uses only credible body parts as direct supervision while inferring unsupported parts from text and visible context.
\item We build K700-M together with a matched evaluation protocol under \textit{Keep-All} and \textit{Discard}, and show that, within this partial-observability setting, selective part-level supervision consistently outperforms both clip-level alternatives and scales better with more web data.
\end{itemize}

\section{Related work}
\label{sec:related}

\subsection{Text-to-motion generation}
Early text-to-motion (T2M) work established the basic problem on curated motion datasets through sequence-to-sequence, GAN, and VAE formulations, including Text2Action~\cite{text2action}, Language2Pose~\cite{language2pose}, Action2Motion~\cite{humanact12}, ACTOR~\cite{actor}, TEMOS~\cite{temos}, TM2T~\cite{tm2t}, and MotionCLIP~\cite{motionclip}. These methods showed that language can drive plausible 3D motion synthesis, but they were developed under a relatively clean supervision regime in which each training clip provides a reliable full-body target.

Recent progress has mostly come from stronger generative backbones. Diffusion-based approaches such as MotionDiffuse~\cite{motiondiffuse}, MDM~\cite{mdm}, MLD~\cite{mld}, ReMoDiffuse~\cite{remodiffuse}, and Make-An-Animation~\cite{makeananimation} improve motion quality and text alignment by modeling motion as a continuous denoising process. Discrete-token and language-style methods instead quantize motion and use autoregressive or masked prediction, as in T2M-GPT~\cite{t2mgpt}, MotionGPT~\cite{motiongpt}, and MoMask~\cite{momask}. More recent studies also explore part coordination, fine-grained control, and streaming generation, including ParCo~\cite{parco}, FineMoGen~\cite{finemogen}, MotionStreamer~\cite{motionstreamer}, the recent rethinking study on masked autoregression and diffusion~\cite{rethinking}, and PlanMoGPT~\cite{planmogpt}. ParCo and FineMoGen are the closest to our setting because explicit part streams make partial supervision expressible: without a part-wise representation, one cannot trust the visible arms while masking out unsupported legs in the same clip. Prior methods still assume those part targets are reliable once the sequence is given and use part structure mainly for coordination or control. RoPAR uses the same decomposition for a different purpose: deciding which targets are allowed to contribute to the loss.

\subsection{Learning from web-scale videos}
To overcome the scale limits of MoCap corpora such as AMASS~\cite{amass} and HumanML3D~\cite{humanml3d}, recent work increasingly turns to internet video as a data source. Motion-X~\cite{motion-x} provides a large expressive motion dataset, while more recent vision-language-motion pipelines such as Being-M0~\cite{being-m0}, Being-M0.5~\cite{being-m0.5}, Go to Zero~\cite{gotozero}, and HY-Motion~\cite{hunyuan-mo} push toward million-scale motion learning and more controllable generation. These works broaden action coverage and visual variety beyond classical motion-capture datasets. Most of them focus on data scale, model scale, or controllability. They spend much less attention on deciding what supervision to trust under partial observability.

Their success also depends on advances in monocular 3D human reconstruction, and these pipelines typically use the recovered full body as the training target. Earlier systems such as SPIN~\cite{spin}, VIBE~\cite{vibe}, and PARE~\cite{pare} improved mesh recovery through optimization in the loop, temporal modeling, and part attention. More recent in-the-wild methods such as SLAHMR~\cite{slahmr}, Humans in 4D~\cite{humans4d}, WHAM~\cite{wham}, WHAC~\cite{whac}, and GVHMR~\cite{gvhmr} better recover camera motion, global trajectory, and world grounding from unconstrained video, while RoHM~\cite{rohm} improves robustness through diffusion-based reconstruction. We do not revisit HMR accuracy or million-scale model design here. We focus on a narrower question inside those pipelines: how should a T2M model learn when different body parts have different levels of visual support?

\subsection{Part-aware modeling and imperfect supervision}
Part structure has already proved useful for motion generation and control. ParCo~\cite{parco} coordinates multiple body parts during decoding, FineMoGen~\cite{finemogen} studies fine-grained spatio-temporal motion control, and TEACH~\cite{teach} together with earlier compositional animation work~\cite{compositionalanim} explores how complex instructions can be assembled from simpler motion components. These studies show that part-aware representations help semantic grounding, controllability, and local motion quality. They are also the closest precursors to our setting, because only a part-decomposed model can naturally apply supervision to one body region while withholding it from another. The difference is that earlier part-aware T2M work still assumes each part target is trustworthy once the sequence is given, whereas RoPAR uses visibility to decide which parts are supervised directly and which parts must be inferred.

Our setting is additionally related to learning with imperfect supervision. Robust learning under noisy labels has been studied through peer teaching, sample filtering, and semi-supervised correction, with representative methods such as Co-teaching~\cite{coteaching}, DivideMix~\cite{dividemix}, and noisy-label surveys~\cite{noisylabelsurvey}. However, these methods usually treat reliability at the example level: a sample is trusted, reweighted, or discarded as a whole. Web-mined motion is more structured. Within the same clip, one arm may be well observed while the legs are weakly supported, or vice versa. RoPAR therefore combines part-aware modeling with a reliability-aware training rule: non-visible parts are masked out of direct supervision, and the model learns completion from text together with the remaining visible evidence. This is the key difference between our setting and prior part-aware T2M or generic noisy-label learning.

\section{Method}
\label{sec:method}

\begin{figure*}[t]
\centering
\includegraphics[width=\linewidth]{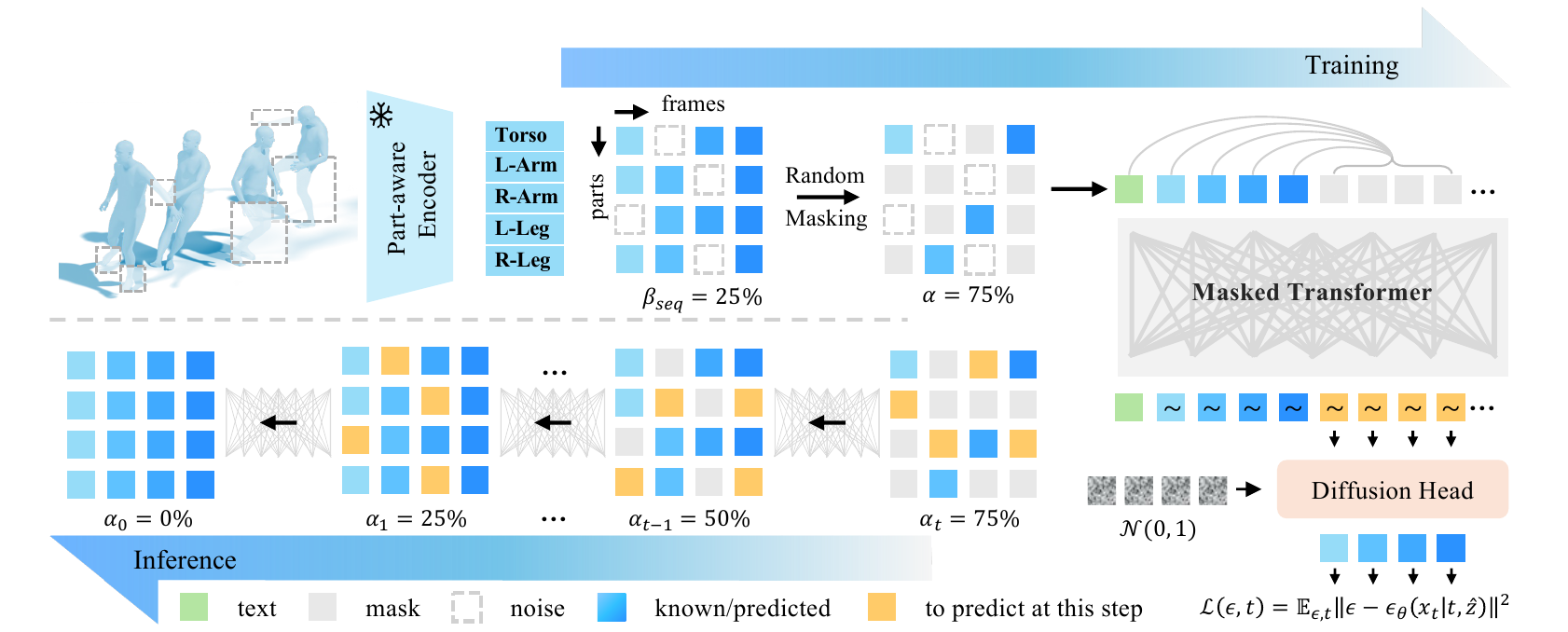}
    \caption{Overview of RoPAR. We first estimate which body parts are credible from 2D pose confidence, then encode only those parts into a part-aware latent grid. During training, non-credible tokens are always masked, a subset of credible tokens is additionally masked, and the generator predicts the missing tokens from text and visible context. A latent diffusion head refines the completed sequence before decoding.}
\label{fig:pmar}
\end{figure*}

Given a text prompt $T$ and a web-mined motion sequence $\mathcal{M} = \{m_i\}_{i=1}^N$ recovered by an HMR system, our goal is to train a full-body generator even though some body parts in $\mathcal{M}$ are not actually observed in the source video. RoPAR addresses this in three steps: (1) estimate a part-level observability mask from 2D pose confidence, (2) learn a part-aware latent space from only credible motion segments, and (3) train a visibility-guided masked generator that predicts missing part tokens from text and visible context.

The key issue is that masked tokens have two different meanings in our setting. Some credible tokens are deliberately masked so that the model learns to predict them. Other tokens correspond to body parts whose HMR labels are weakly supported by the source video; these tokens are hidden from the input because they are unreliable, but they never serve as prediction targets. RoPAR therefore separates conditioning from supervision: the generator learns from text and the remaining credible evidence, while unreliable parts are excluded from direct supervision.

\subsection{Visibility-aware part decomposition}
\label{sec:training}

This step converts image evidence into a part-level credibility signal for the recovered 3D motion. We partition the human skeleton into five kinematic parts $P = \{\text{Torso}, \text{L-Arm}, \text{R-Arm}, \text{L-Leg}, \text{R-Leg}\}$. The 3D motion used for training still comes from the HMR reconstruction $\mathcal{M}$, while ViTPose~\cite{xu2022vitpose} is run on the original RGB frames only to assess whether each body part is sufficiently supported by 2D evidence. Specifically, ViTPose provides a 2D confidence score $C_{i,j} \in [0,1]$ for each joint $j$ at frame $i$, and we define the observability mask of part $p$ at frame $i$ as
\begin{equation}
\small
\mathcal{O}_{i}^{p}=
\begin{cases}
1 & \text{if } \frac{1}{|J_p|}\sum_{j \in J_p} C_{i,j} > \tau,\\
0 & \text{otherwise},
\end{cases}
\label{eq:visibility}
\end{equation}
where $J_p$ is the joint set of part $p$ and $\tau$ is the visibility threshold.

If $\mathcal{O}_{i}^{p}=1$, the part is treated as \emph{credible}. If $\mathcal{O}_{i}^{p}=0$, the HMR output for that part is kept in the sequence for bookkeeping but is not trusted as supervision. This distinction lets us keep partially useful clips. It avoids discarding entire sequences.

\subsection{Part-aware latent modeling}
\label{sec:P-VAE}

We encode each part into a continuous latent token with a part-aware variational autoencoder (P-VAE). For part $p$, the input feature $m_i^p$ contains local joint positions, rotations, and velocities. We also append the root linear velocities $(r^x, r^z)$ and angular velocity $r^a$ so that each part remains grounded in the global body trajectory. The encoder produces a latent token $z_i^p = E_\phi(m_i^p)$, and the decoder reconstructs $\hat{m}_i^p = D_\psi(z_i^p)$.

The key design choice is that the reconstruction loss is applied only to credible parts:
\begin{equation}
\small
\mathcal{L}_{\text{P-VAE}} =
\sum_{i=1}^{N}\sum_{p \in P}
\mathcal{O}_{i}^{p}
\left(
\|m_i^p - \hat{m}_i^p\|_2^2
\;+\;
\lambda_{\text{kl}} \, \mathrm{KL}\big(q_\phi(z_i^p \mid m_i^p)\,\|\,\mathcal{N}(0,I)\big)
\right).
\label{eq:pvae}
\end{equation}
When $\mathcal{O}_{i}^{p}=0$, the part contributes no reconstruction or KL gradient. As a result, the latent space is shaped by observed dynamics. HMR completions do not drive it. We share the encoder and decoder parameters across body parts, which keeps the latent representation comparable across limbs and reduces overfitting to any single part.

After encoding, each motion sequence becomes a latent grid $\mathbf{Z}=\{z_i^p \mid i=1,\ldots,N,\; p \in P\}$ with one token per part per frame.

This tokenization has two practical effects. First, it localizes missing supervision: if only one arm is unreliable, the rest of the sequence is still usable. Second, it makes the later masking step semantically meaningful, because each missing token corresponds to a specific body part at a specific time. The masking pattern is no longer an arbitrary latent chunk.

\begin{table*}[t]
    \centering
    \caption{Main K700-M results under two standard web-data policies. \textit{Keep-All} retains the full recovered source pool and trains on the recovered full-body targets, while \textit{Discard} removes any clip that contains a non-visible body part before training. Best results are in bold and second-best results are underlined. The $\Delta$ row reports RoPAR's absolute improvement over the second-best method in each metric, with positive values always indicating better performance.}
    \label{tab:main_results}
    \begingroup
    \setlength{\tabcolsep}{4pt}
    \begin{tabular}{lcccccccccccc}
    \toprule
    \multirow{2}{*}{Methods} & \multicolumn{6}{c}{\textit{Keep-All} ($\sim$200k)} & \multicolumn{6}{c}{\textit{Discard} ($\sim$50k)} \\
    \cmidrule(r){2-7} \cmidrule(l){8-13}
    & FID $\downarrow$ & R1 $\uparrow$ & R2 $\uparrow$ & R3 $\uparrow$ & Div. $\uparrow$ & MM-D $\downarrow$ & FID $\downarrow$ & R1 $\uparrow$ & R2 $\uparrow$ & R3 $\uparrow$ & Div. $\uparrow$ & MM-D $\downarrow$ \\
    \midrule
    MDM~\cite{mdm} & 37.812 & 0.487 & 0.651 & 0.770 & 6.421 & 17.986 & 35.274 & 0.481 & 0.644 & 0.764 & 6.083 & 18.214 \\
    MotionGPT~\cite{motiongpt} & 30.954 & 0.525 & 0.683 & 0.793 & 6.884 & 17.259 & 31.406 & 0.519 & 0.676 & 0.786 & 6.512 & 17.541 \\
    MoMask~\cite{momask} & \underline{24.334} & 0.585 & \underline{0.761} & 0.834 & 7.551 & 15.827 & \underline{25.118} & 0.582 & 0.754 & 0.828 & \underline{7.134} & 15.968 \\
    MotionStreamer~\cite{motionstreamer} & 25.231 & \underline{0.591} & 0.758 & \underline{0.836} & \underline{7.902} & \underline{15.459} & 25.964 & \underline{0.599} & \underline{0.765} & \underline{0.839} & 6.918 & \underline{15.881} \\
    \midrule
    RoPAR (ours) & \textbf{19.682} & \textbf{0.711} & \textbf{0.864} & \textbf{0.891} & \textbf{9.456} & \textbf{13.612} & \textbf{21.384} & \textbf{0.683} & \textbf{0.842} & \textbf{0.881} & \textbf{8.731} & \textbf{13.948} \\
    $\Delta$ & +4.652 & +0.120 & +0.103 & +0.055 & +1.554 & +1.847 & +3.734 & +0.084 & +0.077 & +0.042 & +1.597 & +1.933 \\
    \bottomrule
    \end{tabular}
    \endgroup
\end{table*}

\subsection{Visibility-guided masked generation}

This stage trains a text-conditioned generator to complete a part-by-time latent grid while avoiding supervision from unreliable parts. The generator receives the latent grid $\mathbf{Z}$ and the text prompt $T$. Non-credible tokens are always masked in the input because their recovered values should not be used as ordinary targets. To keep the learning difficulty comparable across sequences with different visibility ratios, we additionally mask a subset of credible tokens.

Let $\beta_{\text{m}}$ be the fraction of non-credible tokens with $\mathcal{O}_{i}^{p}=0$ in a sequence, and let $\alpha$ be the desired total masking ratio. For credible tokens, we sample a Bernoulli mask with probability
\begin{equation}
\small
\gamma=
\begin{cases}
\frac{\alpha-\beta_{\text{m}}}{1-\beta_{\text{m}}}, & \text{if } \beta_{\text{m}} < \alpha \text{ and } \beta_{\text{m}} < 1,\\
0, & \text{otherwise}.
\end{cases}
\end{equation}
When the non-credible fraction already exceeds the target masking ratio, we therefore set $\gamma=0$ and do not additionally mask credible tokens. Sequences with no credible tokens contribute no term to $\mathcal{L}_{\text{gen}}$.
The final input token $\hat{z}_i^p$ is
\begin{equation}
\small
\hat{z}_i^p=
\begin{cases}
[M] & \text{if } \mathcal{O}_{i}^{p}=0,\\
[M] & \text{if } \mathcal{O}_{i}^{p}=1 \text{ and } r_i^p < \gamma,\\
z_i^p & \text{if } \mathcal{O}_{i}^{p}=1 \text{ and } r_i^p \ge \gamma,
\end{cases}
\label{eq:masking_logic}
\end{equation}
where $r_i^p \sim U(0,1)$ and $[M]$ is a learnable mask token.

The text-conditioned transformer $G_\theta$ predicts the masked credible tokens:
\begin{equation}
\small
\mathcal{L}_{\text{gen}}=
\sum_{i=1}^{N}\sum_{p \in P}
\mathbf{1}\!\left[\mathcal{O}_{i}^{p}=1 \wedge \hat{z}_i^p=[M]\right]
\left\|
z_i^p - G_\theta(\hat{\mathbf{Z}}, T)_{i,p}
\right\|_1.
\label{eq:gen}
\end{equation}
Tokens with $\mathcal{O}_{i}^{p}=0$ are therefore masked in the input but excluded from the training target. They create realistic missing-context patterns, and the model must infer them from text together with the remaining credible tokens. It is never trained to imitate weakly supported labels. In a large corpus, each body part appears as a credible target often enough that the model can learn these completions from text and cross-part context.

This corpus-level coverage matters because RoPAR does not require every clip to contain enough evidence to supervise a full motion. It is designed for datasets in which useful evidence is distributed unevenly across clips. A running action may expose the legs clearly in one clip and the upper body clearly in another. When that variation exists at scale, the model can aggregate these partial observations into a shared text-conditioned motion prior.

The transformer output is a coarse full-body latent grid. We refine it with a lightweight latent diffusion head:
\begin{equation}
\mathcal{L}_{\text{diff}}=
\mathbb{E}_{x_0,t,\epsilon}
\left\|
\epsilon - \epsilon_{\theta}(x_t,t,G_\theta(\hat{\mathbf{Z}},T))
\right\|_2^2,
\end{equation}
where $x_t$ is the noisy version of latent target $x_0$ at diffusion step $t$. This stage improves temporal smoothness and coordination at part boundaries.

The final training objective combines the three components:
\begin{equation}
\mathcal{L}=
\mathcal{L}_{\text{P-VAE}}
\;+\;
\lambda_{\text{gen}}\mathcal{L}_{\text{gen}}
\;+\;
\lambda_{\text{diff}}\mathcal{L}_{\text{diff}}.
\end{equation}
In practice, the three terms play different roles: $\mathcal{L}_{\text{P-VAE}}$ shapes the latent space, $\mathcal{L}_{\text{gen}}$ learns text-conditioned completion from partial observations, and $\mathcal{L}_{\text{diff}}$ sharpens the final latent trajectory.

\paragraph{Training.}
Training follows the same logic end to end. For each web clip, we first recover a 3D motion sequence with HMR and derive part-level observability masks from ViTPose confidence on the original frames. The part streams are then encoded by the shared P-VAE, but reconstruction and KL terms are applied only on credible parts. We next construct the masked latent grid by always masking non-credible tokens and randomly masking additional credible tokens. The transformer predicts only the masked credible targets from text and the remaining visible context, and the diffusion head refines the completed latent trajectory before parameter updates with the joint objective above. This training procedure preserves the structured partial observations in each clip while preventing unreliable parts from acting as direct supervision.

\subsection{Inference}

At test time, no visibility mask is available because the model receives only text. We therefore initialize a fully masked part-by-time latent grid and perform iterative masked generation. At each iteration, the transformer predicts the remaining masked locations, commits the most confident subset of part-by-time tokens, and feeds the updated grid back into the next round. Early iterations establish the coarse action pattern, while later iterations focus on lower-uncertainty local details. Once all latent locations are populated, the diffusion head refines temporal continuity and cross-part coordination, and the shared P-VAE decoder reconstructs the final full-body motion sequence.

This inference setup is consistent with training. During learning, the model repeatedly completes latent grids in which some part-by-time tokens are missing because the corresponding body parts are unobservable or intentionally masked. Inference is simply the limiting case in which all locations are initially missing. Visibility patterns vary across the corpus: legs are clearly observed in some clips, arms in others, and torso motion in many. Under that condition, the model can aggregate complementary partial evidence into a shared text-conditioned motion prior. Each individual clip no longer has to provide full-body supervision.

\section{Experiments}
\label{sec:experiments}

\subsection{Experimental setup}
\label{sec:exp_settings}

\textbf{Datasets.}
We evaluate on both noisy web data K700-M, which is our web-scale dataset and evaluation testbed built from Kinetics-700~\cite{k700}. We process the videos with WHAM~\cite{wham} and obtain 198,627 motion sequences. More than 70\% of these sequences contain at least one body part that is not visible in the source frames. We use Gemini~\cite{gemini} to generate clip-level descriptions and apply the same post-processing pipeline to all clips. For K700-M, we fix a train/validation/test protocol, retrain the standard text-motion evaluator on the K700-M training split, and use the same captions, split, and evaluator for all compared methods. We therefore interpret K700-M primarily as a controlled in-domain comparison under a shared protocol.

\textbf{Implementation details.}
For generation, we train all models with AdamW using learning rate $10^{-4}$ and $\beta=(0.9, 0.99)$. Training uses a 10k-step warm-up schedule, gradient clipping at 1.0, and 50k optimization steps. For evaluation on K700-M, we retrain one shared text-motion evaluator following TMR~\cite{tmr} on the training split using AdamW with learning rate $10^{-4}$, batch size 256, and 100 epochs, and we use this same evaluator for every compared method. The evaluator uses a 128-dimensional motion latent and a 640-dimensional text latent, with 4 transformer layers, 4 attention heads, dropout 0.1, and feed-forward width 1024. Across all experiments, motion is decomposed into the same five body parts as in Sec.~\ref{sec:training}. The credibility threshold $\tau$ is fixed once by validation-time calibration on a held-out subset and then shared across all experiments.

\begin{figure*}[t]
\centering
\includegraphics[width=\linewidth]{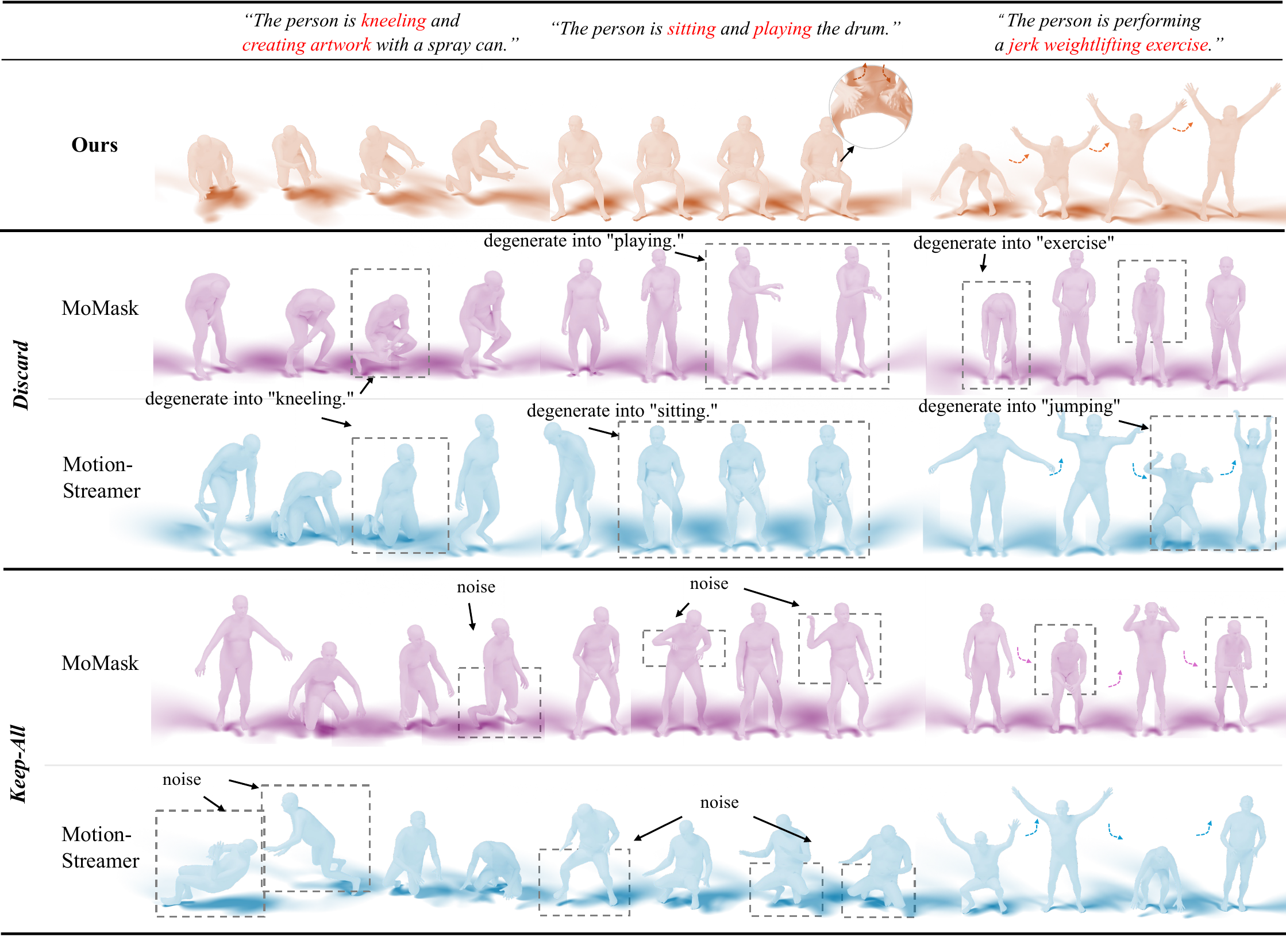}
\caption{Qualitative comparison on K700-M under the \textit{Keep-All} and \textit{Discard} data policies. RoPAR is compared with MoMask and MotionStreamer on prompts where only part of the body is directly observed.}
\label{fig:cases_new}
\end{figure*}

\noindent\textbf{Metrics.}
We report the standard T2M metrics: Fr\'echet Inception Distance (FID), R-Precision (R1/R2/R3), Diversity, and multimodal distance (MM-D). Lower FID and MM-D are better, while higher R-Precision and Diversity are better.

\subsection{Web-data policies}

Our first experimental question is how a full-body T2M system should use web-recovered motion when some body parts are not directly observed. Existing T2M systems are built to align text with a single full-body motion target, whereas many K700-M clips contain recovered motions that are only partly credible. Before comparing architectures, we therefore have to decide how these recovered motions become supervision.

We study two standard web-data policies. In \textit{Keep-All}, we retain the full recovered source pool and train on the HMR output as if the recovered full-body target were trustworthy everywhere. This policy preserves data scale and semantic coverage, but it also exposes the model to unsupported joints. In \textit{Discard}, we remove any clip that contains a non-visible body part and keep only the complete-visibility subset for training. This yields cleaner full-body targets, but reduces the effective training set to roughly one quarter of the original source pool and removes many hard web cases.

These are the only practical policies available to earlier full-body T2M systems on web data. If the model assumes a trustworthy full-body target, it must either trust the recovered full body or train only on clips that remain fully visible. This is also why part-aware models such as ParCo and FineMoGen are the closest prior architectures: only a part-decomposed representation can express supervision for one region of the body while withholding it from another. RoPAR uses that same structure differently. It uses parts to decide where supervision is allowed. We therefore report all K700-M comparisons under two matched settings: a smaller but cleaner setting produced by \textit{Discard}, and a larger but noisier setting produced by \textit{Keep-All}. Table~\ref{tab:main_results} is an end-to-end comparison under those two web-data policies, while the two RoPAR entries also give a direct data-policy ablation of whether keeping the full web source pool helps once unsupported joints are removed from direct supervision.

\subsection{Comparison with baselines}

We compare RoPAR with four strong T2M baselines. MDM~\cite{mdm} is a representative diffusion baseline, while MotionGPT~\cite{motiongpt}, MoMask~\cite{momask}, and MotionStreamer~\cite{motionstreamer} represent language-style motion modeling, masked generation, and causal latent generation, respectively. FineMoGen~\cite{finemogen} and ParCo~\cite{parco} matter especially because they expose explicit body-part structure, but they are more informative in a focused partial-supervision comparison than in the generic baseline table. We therefore use Table~\ref{tab:main_results} for the standard matched K700-M baselines, and later add a dedicated part-aware comparison in Table~\ref{tab:part_partial_compare}. In each K700-M setting of Table~\ref{tab:main_results}, the baselines are retrained on the same selected motion-text pairs as RoPAR and evaluated with the same protocol.

Table~\ref{tab:main_results} reports results under the two matched K700-M settings introduced above: a smaller but cleaner \textit{Discard} setting, and a larger but noisier \textit{Keep-All} setting. Within each column, all methods use the same motion-text pairs, captions, split, and retrained K700-M evaluator. The differences come from training architecture and supervision policy. They do not come from a method-specific scoring backend. Because RoPAR is evaluated under both settings, the gap between its two entries is also a direct ablation of whether retaining the full web source pool helps our model.

The \textit{Keep-All} column is where RoPAR matters most. It outperforms all strong baselines across every reported metric, reducing the best baseline FID from 24.33 to 19.68, improving the best baseline R@1 from 0.591 to 0.711, and increasing Diversity from 7.55--7.90 to 9.46. This is the web-data regime we care about most. Keeping the full source pool helps when the model can ignore unsupported joints. It fails when every reconstructed joint is treated as ground truth.

The \textit{Discard} column makes a different point. Cleaner targets alone still leave a large gap. Removing incomplete clips gives the baselines better full-body supervision. It also shrinks the effective training pool to roughly one quarter of the original web source. Even in this cleaner regime, RoPAR remains strongest, reaching 21.38 FID, 0.683 in R@1, and 8.73 in Diversity. RoPAR is also worse under \textit{Discard} than under \textit{Keep-All}, which means that the removed clips still contain useful supervision once unsupported joints stop acting as targets. The practical takeaway is clear: keep the hard web clips and learn only from the parts that are actually supported by image evidence.

\begin{table}[t]
\centering
\caption{Part-aware comparisons on K700-M in the scale-preserving \textit{Keep-All} regime. We adapt ParCo and FineMoGen to the same credible-part loss masking rule and include a RoPAR ablation without credible-only supervision to isolate the value of selective part-level supervision.}
\label{tab:part_partial_compare}
\begingroup
\setlength{\tabcolsep}{3.5pt}
\begin{tabular}{lccccc}
\toprule
Methods & FID $\downarrow$ & R1 $\uparrow$ & R3 $\uparrow$ & Div. $\uparrow$ & MM-D $\downarrow$ \\
\midrule
ParCo w/ P-Sup. & 23.914 & 0.641 & 0.854 & 8.742 & 14.926 \\
FineMoGen w/ P-Sup. & 21.876 & 0.669 & 0.874 & 8.981 & 14.287 \\
\midrule

RoPAR (ours) & \textbf{19.682} & \textbf{0.711} & \textbf{0.891} & \textbf{9.456} & \textbf{13.612} \\
RoPAR w/o P-Sup. & 32.418 & 0.558 & 0.767 & 7.934 & 15.472 \\
\bottomrule
\end{tabular}
\endgroup
\end{table}

Only part-aware backbones can directly express clip-internal selective supervision, so we adapt ParCo and FineMoGen to the same credible-part masking rule and evaluate them in the scale-preserving \textit{Keep-All} regime. We also include a RoPAR ablation without credible-only supervision to separate the benefit of selective supervision from the rest of the architecture. Table~\ref{tab:part_partial_compare} shows that both part-aware baselines become much stronger once they are trained with partial supervision, which confirms that part decomposition is the right interface for this setting. Removing credible-only supervision from RoPAR also causes a clear drop across all reported metrics, while the full RoPAR model remains best overall. The result points to two ingredients at once: the part-aware architecture matters, and so does using credibility throughout latent learning, visibility-guided masking, and diffusion refinement.

As a fully observed control, we also evaluate RoPAR on HumanML3D, where all training parts are credible and the partial-supervision issue is absent. RoPAR remains competitive there and preserves strong retrieval-style alignment, which suggests that the architecture is not useful only in the target web-partial regime.

\begin{table}[tb]
\centering
\caption{Ablations on K700-M. The top block evaluates P-VAE reconstruction quality, and the bottom block evaluates the full generator.}
\label{tab:ablation}
\begin{tabular}{lcc}
\toprule
Reconstruction & FID $\downarrow$ & MPJPE $\downarrow$ \\
\midrule
P-VAE & \textbf{0.21} & \textbf{6.95} \\
w/o part-wise decomposition & 1.86 & 19.83 \\
w/o shared parameter & 0.89 & 9.82 \\
\midrule
Generation & FID $\downarrow$ & R1 $\uparrow$ \\
\midrule
RoPAR & \textbf{19.68} & \textbf{0.71} \\
w/o part-wise decomposition & 21.36 & 0.58 \\
w/o diffusion head & 71.92 & 0.41 \\
\bottomrule
\end{tabular}
\end{table}

\subsection{Ablations}

\noindent\textbf{Model components.}
This experiment tests whether each design choice is tied to the partial-supervision setting or is simply an isolated engineering trick. Table~\ref{tab:ablation} validates both the P-VAE and the generator. Replacing the part-aware VAE with a full-body VAE increases reconstruction error substantially, which shows that part-level encoding is important even before generation. Removing shared parameters also degrades reconstruction quality, suggesting that a common latent geometry across parts is helpful.

The bottom half of Table~\ref{tab:ablation} shows that the gains do not come from visibility filtering alone. Removing part-wise decomposition hurts both FID and R@1, because the model loses the explicit structure needed to separate credible and non-credible supervision. Removing the diffusion head causes a much larger drop, which indicates that latent refinement matters once the coarse sequence has been assembled from partial evidence.

Taken together, the ablations show that explicit part decomposition is necessary for partial supervision, and that the default five-part split provides a good trade-off between locality and coordination.

\begin{figure*}[t]
    \centering
    \includegraphics[width=\linewidth]{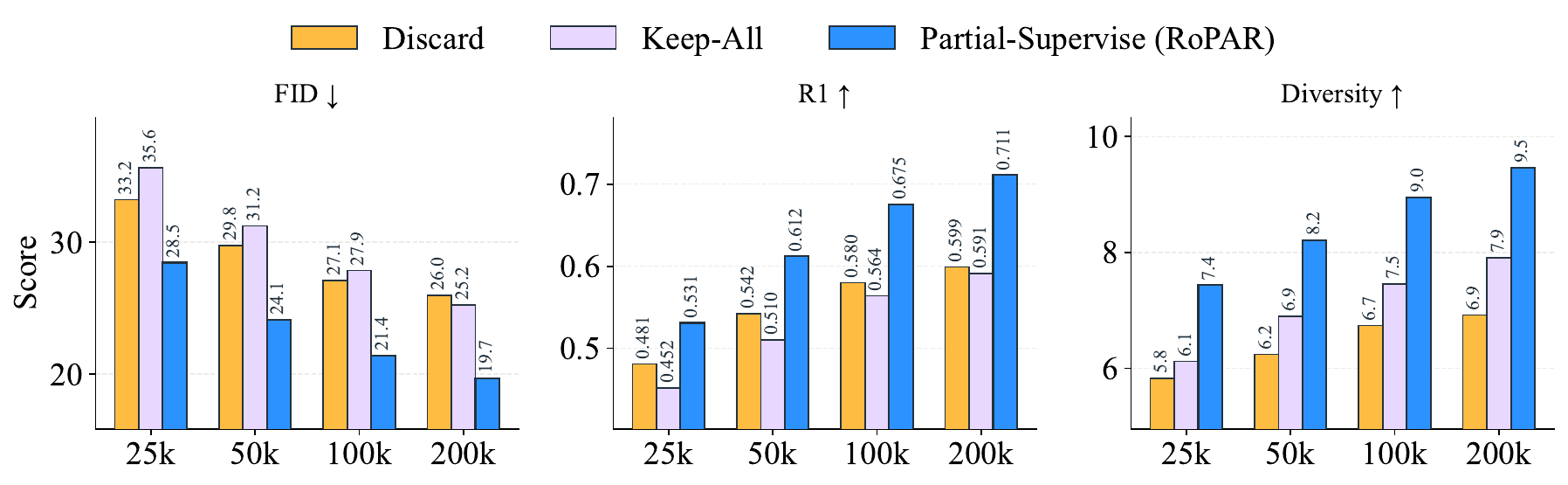}
    \caption{Scaling behavior under unreliable supervision across original recovered K700-M source pools of 25k, 50k, 100k, and 200k clips. The x-axis is indexed by the recovered source pool before policy-specific filtering; under \textit{Discard}, the effective retained set is much smaller because clips with any non-visible body part are removed before training. RoPAR avoids the quality-coverage trade-off between \textit{Keep-All} and \textit{Discard} and achieves the best FID, R1, and Diversity at all four scales.}
    \label{fig:scaling_bar}
\end{figure*}

\noindent\textbf{Scaling behavior.}
Figure~\ref{fig:scaling_bar} extends the comparison to recovered source pools of 25k, 50k, 100k, and 200k clips. All runs use the same source pools, captions, and evaluator, while \textit{Discard} and \textit{Keep-All} also share the MotionStreamer backbone and optimization settings. \textit{Discard} provides cleaner targets and can improve FID and R1 at smaller scales, but filtering hard clips narrows coverage and consistently lowers Diversity. \textit{Keep-All} preserves that coverage yet learns from unsupported joints, which hurts FID and R1. RoPAR avoids both failure modes and achieves the best FID, R1, and Diversity at every scale. The 200k endpoints agree with Table~\ref{tab:main_results}: larger web pools help only when unsupported joints are excluded from direct supervision.

\subsection{Qualitative results}
\label{sssec:ExpQualitative}

Figure~\ref{fig:cases_new} connects the numerical gains to case-level behavior. MoMask usually captures the action category but is less stable in fine part coordination, especially for asymmetric arm use, seated balance, and grounded hand contact. MotionStreamer is more fluent, yet under \textit{Keep-All} it inherits artifacts from unsupported recovered joints: the ``spray can'' motion shows local limb noise, ``playing the drum'' drifts toward an implausible seated pose, and weightlifting loses left-right consistency and temporal smoothness. These errors reflect the same supervision mismatch in different regions of the body: plausible HMR completions are treated as observed targets even when the source frames do not support them.

Under \textit{Discard}, MotionStreamer sees cleaner targets but far fewer hard partial-visibility examples. Local noise can decrease, yet the reduced training coverage makes outputs more conservative and removes the structure needed for seated, crouched, or asymmetric actions. RoPAR avoids both failure modes by retaining these clips without treating unsupported joints as targets. This lets it learn from visible local dynamics across the full web pool while ignoring weak reconstructed labels, preserving asymmetric cues, contact geometry, and smooth part-to-part transitions more consistently across all three examples. The contrast mirrors the quantitative trade-off: \textit{Keep-All} preserves action coverage but propagates unsupported completions, whereas \textit{Discard} suppresses those labels at the cost of rare poses and interactions. Selective part supervision retains the useful side of both policies.

\section{Conclusion}
\label{sec:conclusion}

We study text-to-motion learning from partially observed web motion. RoPAR keeps the full source pool but supervises only image-supported body parts through visibility-aware decomposition, part-aware latent modeling, and masked generation. On K700-M, this selective supervision improves motion quality and text alignment under both \textit{Keep-All} and \textit{Discard}. Its advantage across training scales further shows that partially useful clips are more valuable than either their unsupported joints or wholesale filtering, provided that credibility is modeled inside the objective rather than used only for preprocessing. The HumanML3D control additionally indicates that the architecture remains effective when all parts are credible.

More broadly, the bottleneck in web-scale T2M is not only recovery quality, but also the mismatch between what the video reveals and what the model is asked to imitate. Partial observability is therefore structured missing supervision rather than a binary data-cleaning problem. Aligning the training target with image evidence lets a model use difficult clips without learning prior-based completions as ground truth. The formulation also separates data utility from label completeness: a clip need not provide a trustworthy full body to contribute reliable local dynamics. This perspective may extend to other video-driven 3D generation tasks in which recovered signals combine direct observations with model-based inference.

\noindent\textbf{Limitations.}
RoPAR depends on the calibration of its visibility estimator: false negatives discard useful supervision, while false positives retain weakly supported parts. Its fixed threshold may also require recalibration when the pose estimator or video domain changes. The current study targets single-person videos and does not explicitly address multi-person interaction or severe camera motion. Extending credibility estimation to these cases remains future work.

\bibliographystyle{ACM-Reference-Format}
\bibliography{main}
\end{document}